\documentclass{article}

\PassOptionsToPackage{numbers, compress, sort}{natbib}
\usepackage[final]{neurips_2020}

\usepackage[utf8]{inputenc} % allow utf-8 input
\usepackage[T1]{fontenc}    % use 8-bit T1 fonts
\usepackage{hyperref}       % hyperlinks
\usepackage{url}            % simple URL typesetting
\usepackage{booktabs}       % professional-quality tables
\usepackage{amsfonts}       % blackboard math symbols
\usepackage{nicefrac}       % compact symbols for 1/2, etc.
\usepackage{microtype}      % microtypography
\usepackage{xcolor}
\usepackage{multirow}
\usepackage{bm}
\usepackage{amsmath}
\usepackage{amssymb}
\usepackage{graphicx}
\usepackage{caption}
\usepackage{subcaption}
\usepackage{makecell}
\usepackage{wrapfig}

\title{A Multimodal Framework for the \\ Detection of Hateful Memes}

\author{%
  Phillip Lippe\thanks{The authors contributed equally to this work.} \\ % Correspondence to: \texttt{p.lippe@uva.nl}
  \textsc{QUVA}\\ University of Amsterdam \\ %QUVA \\
  \texttt{p.lippe@uva.nl} \\
   \And
   Nithin Holla$^*$ \\
   University of Amsterdam \\
   \texttt{nithin.holla7@gmail.com} \\
   \And
   Shantanu Chandra$^*$ \\
   ZS\\
   University of Amsterdam \\
   \texttt{shantanu.chandra@zs.com} \\
   \AND
   Santhosh Rajamanickam \\
   Slimmer AI \\
   \texttt{rajamanickamsanthosh@gmail.com} \\
   \And
   Georgios Antoniou \\
   King's College London \\
   \texttt{georgios.antoniou@kcl.ac.uk} \\
   \AND
   Ekaterina Shutova \\
   University of Amsterdam \\
   \texttt{e.shutova@uva.nl} \\
   \And
   Helen Yannakoudakis \\
   King's College London \\
   \texttt{helen.yannakoudakis@kcl.ac.uk} \\
}

\begin{document}

\maketitle

\begin{abstract}

An increasingly common expression of online hate speech is multimodal in nature and comes in the form of \textit{memes}. Designing systems to automatically detect hateful content is of paramount importance if we are to mitigate its undesirable effects on the society at large. The detection of multimodal hate speech is an intrinsically difficult and open problem: memes convey a message using both images and text and, hence, require multimodal reasoning and joint visual and language understanding. In this work, we seek to advance this line of research and develop a multimodal framework for the detection of hateful memes. We improve the performance of existing multimodal approaches beyond simple fine-tuning and, among others, show the effectiveness of upsampling of contrastive examples to encourage multimodality, and ensemble learning based on cross-validation to improve robustness. We furthermore analyze model misclassifications and discuss a number of hypothesis-driven augmentations and their effects on performance, which we hope shall inform future research in the field. Our best approach comprises an ensemble of UNITER-based \citep{chen2019uniter} models and achieves an AUROC score of $80.53$, placing us 4th on Phase 2 of the 2020 \textit{Hateful Memes Challenge} organized by Facebook. 
    
\end{abstract}

\section{Introduction} 

Online abuse is an important societal problem of our time and one that is highly correlated with the rise of social media platforms such as Twitter and Facebook. A large number of Internet users have either encountered or witnessed some form of abusive behaviour online. For instance, it has been reported that $41\%$ of American adults have experienced online harassment \citet{duggan2017men}. Hate speech, a prevalent form of online abuse, has seen a dramatic rise in recent years \cite{hate_speech}. It can be defined as any form of communication that attacks or uses discriminatory language with reference to a person or a group based on their race, gender, religion, etc. \textit{Memes} -- that have recently emerged as popular engagement tools and which, in their usual form, are image macros shared through social media platforms mainly for amusement -- are also being increasingly used to spread hate and/or instigate social unrest, and therefore seem to be a new form of expression of hate speech on online platforms. \textit{Hateful memes} often target certain communities or individuals based for example on religion, gender, race, or physical attributes, by portraying them in a derogatory manner and/or by reinforcing stereotypes. Such memes promote racism \citep{williams-racism} and sexism \citep{drakett-sexist} among other forms of hate speech, threatening social peace and leading to damage at both the individual and societal level \citep{prithvi-adverse}.

\begin{wrapfigure}{r}{0.35\textwidth}
  \centering
  \vspace{-3mm}
  \includegraphics[width=0.35\textwidth]{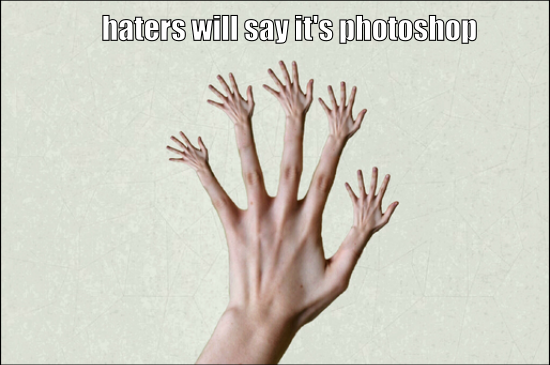}
  \caption{Example (non-hateful) multimodal meme from the HM dataset \cite{kiela2020hateful}. The image is a compilation of assets, including ©Getty Images.}
  \label{fig:intro_example_meme}
  \vspace{-2mm}
\end{wrapfigure}

In this light, developing systems that can automatically detect hateful memes (and hateful content in general) is of paramount importance if we are to mitigate their undesirable effects. However, the detection of multimodal hate speech is an intrinsically difficult and open problem within the joint visual and language (V+L) understanding domain as it requires a holistic understanding of content, where reasoning about image and text is simultaneous. As the example in Figure~\ref{fig:intro_example_meme} shows, it is not sufficient to rely on the text or the image modality individually for correct interpretation of content; rather both modalities should be jointly processed to infer the correct meaning of the meme. Due to the multimodal nature of the problem involving an interplay between the image and the text, existing and state-of-the-art multimodal systems perform rather poorly on the detection of hateful memes \cite{kiela2020hateful}. This highlights the need to further advance multimodal reasoning and understanding for systems to be able to more accurately flag hateful content online. 

With that in mind, Facebook launched the \textit{Hateful Memes Challenge} \cite{kiela2020hateful} as part of the NeurIPS 2020 competition track to encourage further research into multimodal reasoning and the design of systems that can detect hateful memes. Specifically, the task is formulated as a binary classification problem, where a meme can belong to one of two classes -- \textit{hateful} or \textit{not hateful}. The challenge introduces a new Hateful Memes (HM) dataset \citep{kiela2020hateful}, consisting of over $10,000$ memes. To enforce multimodality and reduce accidental biases in system classifications, the dataset includes non hateful, \textit{benign confounders} (i.e., \textit{contrastive} or \textit{counterfactual} examples) for a subset of hateful memes. This makes the task particularly challenging for unimodal systems that only leverage signals from either of the two modalities and do not combine information from both the text and the image. This is corroborated by the performance of the baselines implemented by the organisers of the challenge \cite{kiela2020hateful}, which clearly shows a substantial difference in  performance between unimodal and multimodal systems; albeit the latter still performing rather poorly, and particularly compared to human performance, indicating that a more refined understanding of multiple signals is necessary. 

From their thorough baseline experiments, the challenge organisers concluded that early-fusion multimodal models considerably outperform late-fusion architectures for the task. Building on their work, we experiment with a number of early-fusion multimodal approaches, namely LXMERT \cite{tan2019lxmert}, UNITER \cite{chen2019uniter}, and Oscar \cite{Li2020OscarOA}, for the task of hateful memes detection. We propose various methods for improving model performance beyond simple fine-tuning and, among others, show the effectiveness of upsampling confounders to encourage multimodality, and ensemble learning based on cross-validation to improve model robustness. 
Our best system is an ensemble method based on UNITER that achieves an AUROC score of $80.53$, ranking our team, \textit{Kingsterdam}, 4th on the final phase (Phase 2) of the Hateful Memes Challenge. We conclude with an error analysis of model misclassifications and discuss a number of hypothesis-driven model augmentations based on multi-task learning and their effects on performance, which we hope shall inform future research in the field. To facilitate further research, we make our code publicly available.\footnote{\url{https://github.com/Nithin-Holla/meme_challenge}} 

\section{Related work}

\subsection{Multimodal learning} 

Multimodal representation learning has recently gained traction due to the poor performance of existing (unimodal) models on multimodal tasks such as Visual Question Answering \citep{antol2015vqa,hudson2019gqa} and Visual Reasoning \citep{suhr2018corpus}. These tasks involve V+L understanding and identifying the synergy between the two modalities. Most existing multimodal systems adopt either a late-fusion (LF) or an early-fusion (EF) approach to process the two modalities. Late-fusion methods \citep{kiela-bitransformers,kiela2020hateful} typically utilize unimodal models to process the two signals independently and then combine their features (usually via simple concatenation) before the final classification layer. Early-fusion methods such as MMBT \citep{kiela-bitransformers},  VisualBERT \citep{li2019visualbert}, and ViLBERT \citep{lu2019vilbert}, on the other hand, employ more complex approaches to process the two modalities jointly within the model architecture. UNITER (UNiversal Image-TExt Representation) \cite{chen2019uniter}, LXMERT (Learning Cross-Modality Encoder Representations from Transformers) \cite{tan2019lxmert} and Oscar (Object-Semantics Aligned Pre-training) \cite{Li2020OscarOA} are some of the recent and popular early-fusion multimodal representation learning models which are pretrained on various V, L and V+L tasks such as visual question answering and image captioning, and which we use in our work (discussed in detail in Section \ref{models}). Oscar, the more recent of the three, outperforms LXMERT and UNITER on the VQA \citep{antol2015vqa}, GQA \cite{hudson2019gqa}, and NLVR2 \cite{suhr2018corpus} downstream tasks and achieves state-of-the-art performance on VQA and NLVR2, among others.

\subsection{Multimodal hate speech detection} 
Existing work on hate speech detection has largely relied on text-based features. \citet{mishra2019tackling} discuss emerging trends, resources and challenges, as well as outline the various approaches used in the domain of online abusive language detection. However, there is comparatively little work in the vision or multimodal domain, something which can be attributed to the scarcity of annotated datasets. \citet{gomez2020exploring} introduce MMHS150K, a multimodal dataset of tweets consisting of both image and text and which are manually annotated for hate speech. The authors develop and evaluate three multimodal models -- Feature Concatenation Model (FCM), Spatial Concatenation Model (SCM) and Textual Kernels Model (TKM) -- and conclude that they fail to outperform unimodal counterparts. \citet{hosseinmardi2015detection} discuss the problem of cyberbullying and cyberaggression in Instagram posts and comments and evaluate the performance of a Naive Bayes classifier and a linear SVM classifier using a range of features, such as word n-grams and image categories, as well as metadata such as the number of followers and likes.  
%The text features were n-gram features obtained from the content of comments, while the features extracted from user and media information (indicated as metadata) includes the number of followers, followings, likes, and shared media and features extracted from the image content included image categories. 
%The experiments show that the linear SVM classifier with the multimodal features as input outperformed the Naive Bayes classifier variants. 
%%The authors also suggest that simple metadata features can be used to improve accuracy, but to increase recall, more complex features were needed.

% Notable works in multimodal hate speech include the detection of hateful multimodal tweets using the MMHS150K dataset \cite{gomez2020exploring}, the detection of cyberbullying and cyberaggression in Instagram posts and comments \cite{hosseinmardi2015detection}, and an interpretable multimodal hate speech classification with modalities involving text and socio-cultural information \cite{vijayaraghavaninterpretable}.

With regards to the \textit{Hateful Memes Challenge}, the organizers provide several baseline models \cite{kiela2020hateful} and evaluate their performance using accuracy and AUROC. The baselines comprise unimodal and multimodal systems which are pretrained using diverse methods. Their experiments show that multimodal approaches outperform all the unimodal systems, reiterating the fact that this task requires a holistic understanding of the meme by processing the image and text signals jointly. 

The multimodal approaches include several architectures based on BERT \citep{devlin2018bert} that combine image and text features to get the final prediction. The best-performing multimodal baseline models are ViLBERT \cite{lu2019vilbert} and VisualBERT \cite{li2019visualbert} which are pretrained on the CC (Conceptual Captions) \citep{sharma2018conceptual} and the COCO (Common Objects in Context) \citep{lin2014microsoft} datasets respectively. ViLBERT extends the BERT architecture to a multimodal, two-stream model and incorporates separate transformers for the vision and language domain that interact through co-attentional transformer layers to learn joint representations of images and text. VisualBERT is a single-stream architecture which uses a self-attention mechanism within a layer of the transformer which includes both vision and language inputs.

\section{Dataset} 
\label{data}

The challenge uses the Hateful Memes (HM) dataset \cite{kiela2020hateful} compiled by Facebook AI. It includes the memes (image with text) as well as the meme text separately to facilitate easier processing. The dataset consists of over $10$k memes labeled as \textit{hateful} or \textit{non-hateful} using the definition of hatefulness presented by \citet{kiela2020hateful} for this task -- ``a direct or indirect attack on people based on characteristics, including ethnicity, race, nationality, religion, caste, sex, etc." (see \cite{kiela2020hateful} for the full definition). The dataset also consists of an additional $2$k unlabeled memes that form the test sets of the challenge and which were released in two phases of the competition: Phase 1 (``seen'' test set) and Phase 2 (unseen and final test set used to determine system rankings).

The dataset is designed such that multimodal approaches are essential to perform well on the task. Specifically, for a set of hateful memes, \textit{benign confounders} are devised and included in the dataset. These are defined as the result of minimal transformations to either the image or the text of the meme such that the corresponding label flips from hateful to non-hateful, or vice versa. Hence, there are two types of confounders -- (a) \textit{image confounders}, where the pair of memes have the same image but different text, and (b) \textit{text confounders}, where the pair of memes have the same text but different images. Overall, the dataset comprises five different types of memes: \textit{multimodal hate}, \textit{unimodal hate}, \textit{benign image} and \textit{benign text} confounders, and finally \textit{random non-hateful examples}. We refer the reader to \citet{kiela2020hateful} for more details on the dataset and its construction.

\begin{table}[t] % NeurIPS has captions above tables
    \caption{Distribution of the different types of memes in the Hateful Memes dataset.}
    \label{tab:data-statistics}
    \vspace{1mm}
    \small
    \centering
    \begin{tabular}{lcccccc}
    \toprule
    \textbf{Split} & \makecell{\textbf{Multimodal} \\ \textbf{hate}} & \makecell{\textbf{Unimodal} \\ \textbf{hate}} & \makecell{\textbf{Benign} \\ \textbf{confounders}} & \makecell{\textbf{Random} \\ \textbf{benign}} & \makecell{\textbf{Dynamic adversarial} \\ \textbf{benign confounders}} & \textbf{Total} \\
    \midrule
    Train  & $1300$ & $1750$ & $3200$ & $2250$ & -- & $8500$ \\
    Dev-seen & $200$ & $50$ & $200$ & $50$ & -- & $500$ \\
    Test-seen & $400$ & $100$ & $400$ & $100$ & -- & $1000$ \\
    Dev-unseen & $200$ & -- & $200$ & -- & $140$ & $540$ \\
    Test-unseen & $729$ & -- & $597$ & -- & $674$ & $2000$ \\
    \bottomrule
    \end{tabular}
\end{table}

There is a total of $12,140$ memes, split into train, development (validation) and test sets. The training set contains $8,500$ examples, with $36\%$ hateful and $64\%$ non-hateful memes. The development and test sets are composed of two sets each corresponding to Phase 1 (\textit{dev-seen}, \textit{test-seen}) and Phase 2 (\textit{dev-unseen}, \textit{test-unseen}) of the competition. The \textit{dev-seen} and \textit{test-seen} sets are class-balanced and contain $500$ and $1,000$ examples respectively. In these, the distributions of the five different types of memes are as follows: $40\%$ multimodal hate, $10\%$ unimodal hate, $20\%$ benign text, $20\%$ benign image confounders and the remaining $10\%$ are random non-hateful. The \textit{dev-unseen} set includes multimodal hateful and benign confounders from \textit{dev-seen}, as well as a new set of $140$ dynamic adversarial benign confounders that are chosen such that the VisualBERT model fails to classify them correctly. The \textit{test-unseen} set comprises several new multimodal hate and benign confounders as well a new set of dynamic adversarial benign confounders based on VisualBERT, resulting to a total of $2,000$ examples. The composition of \textit{test-unseen} makes this a challenging set and an appropriate test of multimodality. We provide the details of the distribution of the different types of memes on the various splits of the dataset in Table \ref{tab:data-statistics}.

\section{Models}
\label{models}
Early-fusion architectures were shown to outperform the late-fusion ones for the task \cite{kiela2020hateful}; therefore, we focus on three early-fusion pretrained models, namely LXMERT \cite{tan2019lxmert}, UNITER \cite{chen2019uniter} and Oscar \cite{Li2020OscarOA}.

\subsection{LXMERT}
LXMERT is a large-scale transformer model which first processes the image and text by two independent, unimodal encoders. Another consecutive encoder combines the two unimodal representations via cross-attention modules. The model is pretrained on images from COCO and Visual Genome \cite{krishna2017visual} as well as the image question answering datasets -- VQA v2.0 \cite{Goyal2017MakingTV}, GQA \cite{hudson2019gqa}, and VG-QA \cite{Zhu2016Visual7WGQ}. Specifically, it is pretrained on the tasks of masked language modeling, masked object prediction, cross modality matching, and image question answering. LXMERT has been shown to outperform ViLBERT and VisualBERT on downstream tasks such as visual reasoning tasks.

\subsection{UNITER}
UNITER is an early-fusion transformer model which utilizes self-attention on the joint image and textual input. It encodes visual features from bounding boxes using an image encoder and encodes word tokens using a text encoder into a common embedding space. The image input consists of features from Faster R-CNN \cite{ren2015faster} and $7$-dimensional location features consisting of normalized top / left / bottom / right coordinates, width, height, and area for each of the bounding boxes. The text input consists of word embeddings as well as position embeddings. It is followed by several layers of self-attention. 

The model is pretrained on four tasks, namely masked language/region modeling conditioned on image/text (MLM and MRM), image-text matching (ITM) and word-region alignment (WRA). The large-scale pretraining is performed over four V+L datasets -- COCO, Visual Genome, CC, and SBU Captions \cite{Ordonez2011Im2TextDI}. UNITER has been shown to  outperform baselines including LXMERT on six V+L tasks such as masked language modeling (MLM), ITM and MRFR (Masked Region Feature Regression) among others, across nine datasets. 

\subsection{Oscar}
Oscar is another multimodal transformer model whose input consists of triples containing the word sequence, the set of object tags detected in the image, and the set of image region features. The object tags act as \textit{anchor points} that make learning of semantic alignments between images and texts easier. Faster R-CNN is used to produce bounding box features as well as the set of object tags. Additionally, a $6$-dimensional position vector is included for each of the bounding boxes.

The large-scale pretraining consists of two objectives -- a masked token loss and a contrastive loss. Pretraining is performed on V+L datasets including COCO, CC, SBU, Flickr30k \citep{young-flicker}, VQA \citep{antol2015vqa}, GQA, and VG-QA. Oscar was shown to obtain state-of-the-art results on six downstream tasks such as VQA, VCR and NLVR, among others.

\section{Approach}

In this section, we present the various design aspects of our proposed framework for this task. We begin by describing how we extract image features for this dataset. We then describe how simple learning strategies can help us perform better on multimodal hate and benign confounders. We follow that with detailing the importance of creating an ensemble of models to help the model generalize better, and how this ensemble can be effectively optimized using an evolutionary algorithm (EA) for optimal weighting of model predictions. Finally, we discuss how supplementing the model with additional information from fine-grained object-detection classes may help the model identify the target groups in the image and the meme text.

\subsection{Image feature extraction and base model selection}
\label{51}
All our multimodal architectures -- LXMERT, Oscar and UNITER (Section \ref{models}) -- take both tokenized text as well as processed image features as input. To that effect, we first extract the image features for the HM dataset using a pretrained Faster R-CNN \citep{ren2015faster} backbone. For this, we made sure that we use the corresponding pretrained model checkpoint\footnote{\url{https://github.com/MILVLG/bottom-up-attention.pytorch}} as described in the original papers of these architectures rather than the features provided in the MMF library \cite{mmf} for the dataset and released as part of the competition. This was done to ensure that the input features of the meme images are in the same latent space as those that these multimodal pretrained models were trained on, to avoid erroneous behavior. The tokenized text features were obtained using the standard pretrained BERT tokenizer \cite{devlin2018bert} for the models, inline with their pretraining step. 

Using the text and image features as input, we first fine-tune each of the models on the HM dataset (both the base and large versions of the models) for the supervised task of hateful memes detection using a binary classification objective. Specifically, we optimize for the binary cross-entropy (BCE) loss which is computed as:
\begin{equation}
    \mathcal{L}_{BCE}(\bm{\theta}) = \sum_{i=1}^N - \left[ y_{i} \log f_{\bm{\theta}}(x_{i}) + \left( 1 - y_{i} \right) \log \left( 1 - f_{\bm{\theta}}(x_{i}) \right) \right]
\end{equation}
where $f_{\bm{\theta}}$ is the output of the model, $x_{i}$ is the input and $y_{i}$ is the gold label.

We find that UNITER outperforms LXMERT and Oscar by a large margin for both their base and large versions (see results for base models in Table \ref{tab:results_main}), which could be attributed to its diverse set of pretraining tasks. We also note that the large versions of these architectures lead to poor generalization performance due to more severe overfitting on the training data as a result of their larger set of parameters. Given these observations, we proceed with UNITER-base (hereafter referred to as UNITER) as the base model of our framework and use it for all subsequent experiments. 

Using the pretrained UNITER (base), we also experimented with an additional pre-training / domain-adaptation step on the HM dataset in an attempt to align the model's weights to the new latent space of the memes domain. Specifically, we took the pretrained UNITER-base model and fine-tuned it further on the MLM, ITM and MRFR (within MRM) pretraining tasks using the HM dataset. This ``warm-up'' pretraining phase was then followed by the supervised fine-tuning step on the binary task of hateful memes detection. However, this did not yield any performance improvements and we therefore only apply supervised fine-tuning.

\subsection{Confounder upsampling and loss re-weighting}
\label{sec:cfu_and_pw}

A key characteristic of the HM dataset is the inclusion of benign confounders to counter the possibility of models exploiting unimodal priors rather than learning to reason multimodally. Thus, these sets of memes form an important source of truly multimodal instances that the model can directly leverage for effective learning. 

During our model-benchmarking experiments (Section \ref{51}), we noticed that performance was quite poor on benign confounders and specifically text confounders. The models did not effectively exploit these instances during training such that they can multimodally infer their true, underlying meaning. As a simple strategy to alleviate this problem, we upsample the confounders in the batches during the supervised fine-tuning step of UNITER. As the model performed quite well on the image confounders, we only upsample text confounders. Intuitively, this helps the model to focus on input features of both modalities and subsequently improve its performance. We refer to the confounder upsampling approach using the abbreviation CFU.

Another important characteristic of the data that directly affects learning is the gold label distribution. The HM training set consists of 36\% hateful and 64\% non-hateful memes. To improve the detection of hateful memes, we employ a loss re-weighting strategy during training and weigh the loss of the hateful class higher, subsequently affecting the updates of the model parameters. Thus our new loss function is defined as follows:
\begin{equation}
    \mathcal{L}_{HW}(\bm{\theta}) = \sum_{i=1}^N - \left[ \alpha_{pos} \cdot y_{i} \log f_{\bm{\theta}}(x_{i}) + \alpha_{neg} \cdot \left( 1 - y_{i} \right) \log \left( 1 - f_{\bm{\theta}}(x_{i}) \right) \right]
\end{equation}
where $\alpha_{pos}$ and $\alpha_{neg}$ are the weights for the hateful and not-hateful classes respectively such that $\alpha_{pos} > \alpha_{neg}$ and $\alpha_{pos} + \alpha_{neg} = 1$. 
We refer to this loss re-weighting approach on the hateful class using the abbreviation HW.
%We denote loss re-weighting on the hateful class by the abbreviation HW in our discussions later.

\subsection{Cross-validation ensemble}
\label{sec:crossval}
Multimodal datasets are typically much smaller in size than other visual datasets used to train deep image classification models. Training on different subsets of the rather small training set of the HM dataset %we experienced that training multiple models on slightly different subsets of data 
can lead to considerable variation in model predictions. 
To stabilize the predictions and tackle overfitting, we utilize an ensemble of UNITER models where we combine them using a weighted average of their predictions. The weight of each model is determined by an evolutionary algorithm (EA) that optimizes the weights with respect to the AUROC score of the ensemble on the development set. We denote models trained using cross-validation folds derived from the HM training set by the abbreviation CV.
% The specifics of the EA did not seem to have a significant impact on performance.

\begin{figure}[t!]
    \centering
    \includegraphics[width=\textwidth]{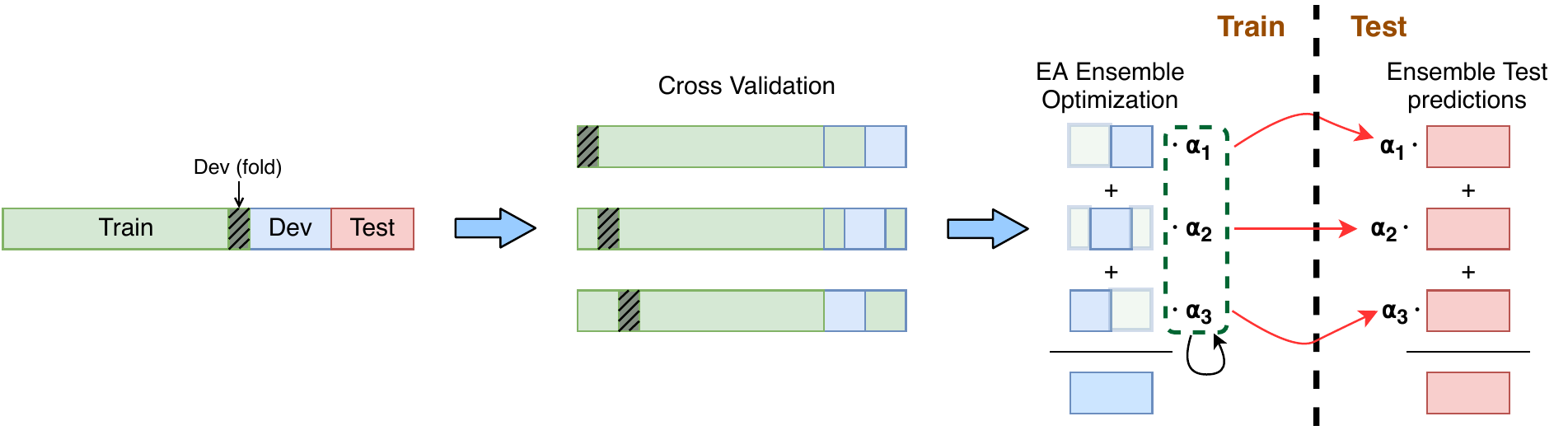}
    \caption{A representation of our cross-validation ensemble optimization process. The dev set (dev-seen; blue) is split into two parts (such that confounder pairs remain together) which are used to augment the CV training and test folds (green). The augmented CV test folds (green+blue) are used for the EA optimization on the right. The final ensemble weights $\alpha_1,...,\alpha_M$, where $M$ is the number of CV folds, are used to combine the CV model predictions and evaluate performance on the final, unseen test set.} % Given a training, development, and test split, we combine the train and dev split to perform cross-validation on. 
    %One split is used for training, the other for testing.
    \label{fig:cross_val}
\end{figure}

The development set contains a good distribution of true multimodal examples which can be valuable during training. To fully utilize the data, exploit as many multimodal examples as possible and test generalization performance of fine-tuned parameters, we employ another cross-validation strategy such that we can also learn from the development set as well as use it for ensemble optimization. 
%Thus, it is important to let the models train on this part of the dataset as well. At the same time, we want to keep the development set separate so that the EA optimization of the ensemble weights is performed on an unseen dataset similar to the test set. 
%To allow us to train on the development set, and yet use it for ensemble optimization, we employ a cross-validation strategy.
Specifically, we split the HM training set into CV folds (implementation details in Section \ref{sec:exp_and_results}) and, in each training fold we include half of the development set (dev-seen), while the other half is added to the test fold. While splitting the development set, we ensure that text confounder pairs (with different labels) remain together. % to increase the confounder amount for training. 
This means that examples that form a confounder pair are never split across the CV training and test sets. The splitting of the development set is otherwise dynamic, i.e., random subsets of the development set are selected for inclusion in the CV folds. 
%included in the CV folds for training and testing for each fold. 
EA optimization of the ensemble weights is now performed on the augmented CV test folds. A visual representation of this cross-validation ensemble optimization process is presented in Figure \ref{fig:cross_val}. We denote models/ensembles trained using the augmented cross-validation folds with the subscript FINAL.
%Note that this \textit{dev-seen} splitting style cross-validation is employed only in the final model with the subscript FINAL.

\subsection{Training with margin ranking loss}

In order to explicitly encourage learning from confounders, we also experiment with training using margin ranking loss as a means of contrastive learning. For every meme $x_{i}$ with label $y_i$, we sample another meme $\tilde{x}_{i}$ with label $\tilde{y}_i$ such that $\tilde{x}_{i}$ is the text confounder of $x_{i}$ if it exists, or a randomly sampled meme from the training set such that $y_{i} \neq \tilde{y}_i$. Training is performed on input pairs $\{ (x_{i}, y_{i}, \tilde{x}_{i}, \tilde{y}_{i}) \}_{i=1}^{N}$ where $N$ is the total number of memes in the training set. The objective is to predict a higher probability score for the meme that is labeled as hateful. The new training loss is then calculated as the weighted sum of binary cross-entropy loss defined earlier and margin ranking (MR) loss:
\begin{equation}
    \mathcal{L}(\bm{\theta}) = \mathcal{L}_{HW}(\bm{\theta}) + \gamma \mathcal{L}_{MR}(\bm{\theta})
\end{equation}
where $\bm{\theta}$ is the set of model parameters and $\gamma$ is a hyperparameter that controls the influence of the MR loss. The BCE loss $\mathcal{L_{HW}}$ is computed as described in Section \ref{sec:cfu_and_pw}, while the MR loss is computed using the input pair as follows:
\begin{equation}
    \mathcal{L}_{MR}(\bm{\theta}) = \sum_{i=1}^N \max \left[ 0, -y_{RANK} \left( f_{\bm{\theta}}(x_{i}) - f_{\bm{\theta}}(\tilde{x}_{i}) \right) + m \right]
\end{equation}
where $m$ is the margin hyperparameter and $y_{RANK} = 1$ if $y_{i}=1$ (hateful) and $\tilde{y}{i} = 0$ (not hateful), and $y_{RANK} = -1$ otherwise. At testing time, evaluation is performed in the standard setting, i.e., without any instance pairing. We refer to this training approach using the abbreviation MRL. 
%denote the model training using margin ranking loss with the abbreviation MRL.

\subsection{Fine-grained YOLO9000 object tags}
\label{yolo}
During qualitative analyses of our models, we find misclassification errors on memes promoting hate towards certain communities or individuals based on religion, gender, and race, among others. We hypothesize that augmenting the model with information that would help identify the target group may lead to more accurate predictions. %help to more accurately identify hate. 
Specifically, we develop a variant of our model that uses YOLO9000 predictions \citep{redmon-yolo9000}. YOLO9000 is a well-known image object detection model that can detect objects from a pre-defined set of $9000$ fine-grained classes. Since not all of these classes are relevant to the HM dataset, we instead use a subset of $97$ classes, such as ``Nigerian'', ``Muslimah'', ``revolver'', and ``amputee'' (see Section \ref{sec:yolo_classes} for more details). For every meme, we take the output classes predicted from the pretrained YOLO9000 model (multi-label classification), filter them to use only our selected set, and provide them as additional input to the model. Concretely, the input now consists of the meme text, YOLO9000 object tags, and bounding box features. We refer to this model using the subscript YOLO.

\section{Experiments and Results}
\label{sec:exp_and_results}

In this section, we provide our implementation details and the hyperparameters of our best-performing models. We then present and discuss the performance of a number of model variants. 

\noindent \paragraph{Hyperparameters} For all our experiments, we use UNITER-base as our main model. We work with a batch size of $16$ with gradient accumulation of $2$ (making the effective batch size $32$) due to memory constraints. We train the models with a learning rate of $3\times10^{-5}$ coupled with the cosine learning rate scheduler using $500$ warmup steps. We optimize the binary cross entropy loss using the Adam optimizer \cite{adam} with a weight decay of $1\times10^{-3}$. During training, we optimize for the AUROC metric with early stopping patience of $5$ for a maximum of $30$ epochs. We limit the maximum text length to $60$ tokens and image bounding box features to $100$ per meme. Furthermore, we upsample the text confounders during training by a factor of $3$ (which we empirically found to perform best), i.e., confounders are $3$ times more likely to be sampled while forming a batch. Additionally, we scale up the loss of the hateful class by a factor of $1.8$, attempting to mititgate the effect of the unbalanced training set distribution. 
We experiment with varying numbers of cross-validation folds; however, our best model is trained under a $15$-fold cross-validation setting. We optimize the CV ensemble weights using an evolutionary algorithm with tournament selection, Gaussian noise mutation and uniform crossover \cite{Miller1995GeneticAT, EAEiben} strategies. Out of a population of $512$ individuals and $100$ generations, we pick the set of ensemble weights that achieve the highest AUROC score on the development set (dev-seen).

\begin{table}[t]
    \caption{AUROC scores of our models on the development set (dev-seen) as well as the Phase 1 and 2 test sets of the challenge.}
    \label{tab:results_main}
    \vspace{1mm}
    \centering
    % \small
    \begin{tabular}{lccc}
        \toprule
        \multirow{2}{*}{\textbf{Model}} & \multicolumn{3}{c}{\textbf{AUROC}} \\ 
                                        & \textbf{Development} & \textbf{Phase 1} & \textbf{Phase 2} \\ \midrule
        ViLBERT CC & 70.07 & 70.03 & -- \\
        VisualBERT COCO & $73.97$ & $71.41$ & -- \\ \midrule
        % LXMERT\textsubscript{MMF} & $67.12$ & -- & --\\
        % UNITER\textsubscript{MMF} &  $71.20$ & $70.60$ & -- \\
        % UNITER\textsubscript{CFU + MMF} &  $73.18$ & $71.98$ & --\\
        UNITER &  $78.04$ & $74.73$ & -- \\
        LXMERT & $72.33$ & -- & -- \\ 
        % LXMERT\textsubscript{3xCONF + 36-FRCNN} & $71.06$ & -- & -- \\ \midrule
        Oscar &  $72.00$ & -- & -- \\ \midrule
        UNITER\textsubscript{CV10} &  $79.81$ & -- & -- \\ 
        UNITER\textsubscript{CV10 + CFU} &  $79.64$ & -- & -- \\
        UNITER\textsubscript{CV10 + CFU + HW} &  $80.01$ & $78.60$ & -- \\
        % UNITER\textsubscript{CV15 + CFU + HW + 42} &  $81.19$ & $78.64$ & -- \\
        UNITER\textsubscript{CV15 + CFU + HW} &  $80.65$ & $79.06$ & -- \\
        UNITER\textsubscript{CV30 + CFU + HW} & $81.36$ & $78.98$ & -- \\
        UNITER\textsubscript{CV15 + CFU + HW + MRL} & $80.44$ & $78.14$ & -- \\
        UNITER\textsubscript{CV15 + CFU + HW + YOLO} & $80.67$ & $78.21$ & -- \\ \midrule
        UNITER\textsubscript{ENSEMBLE 1} & $81.71$ & $79.13$ & $80.33$ \\
        UNITER\textsubscript{ENSEMBLE 2} & $81.76$ & $79.10$ & $80.40$  \\
        UNITER\textsubscript{FINAL} & $77.39$ & $79.07$ & \bm{$80.53$} \\
        \bottomrule
    \end{tabular}
\end{table} 

\paragraph{Results} The results of our experiments are presented in Table~\ref{tab:results_main}, where we show the AUROC on the development set (dev seen), as well as the Phase 1 and Phase 2 test sets (seen and unseen respectively). The number of Phase 1 and Phase 2 submissions were limited to one per day and three in total respectively, and therefore some of the models have missing entries. 

ViLBERT CC and VisualBERT COCO are the two best performing multimodal baselines provided by \citet{kiela2020hateful}. We first note that simple supervised fine-tuning of UNITER using the HM task objective (denoted as UNITER in Table \ref{tab:results_main}) already produces considerable improvement compared to these multimodal baselines. Similar fine-tuning with LXMERT performs worse compared to VisualBERT and UNITER. Oscar, despite having been shown to achieve  state-of-the-art results on several V+L tasks, obtains lower scores compared to UNITER. 

Continuing with UNITER as our best model, we evaluate the effectiveness of a number of variants. 
UNITER\textsubscript{CV10} is trained using 10-fold cross-validation using only the training set and the data split process described in Section \ref{sec:crossval} (first paragraph). The final predictions are obtained using an ensemble optimized using an evolutionary algorithm as described in Section \ref{sec:crossval}. This achieves a slightly higher validation score on dev-seen, indicating that cross-validation training and ensembling can help boost performance compared to simple fine-tuning. When confounder upsampling (CFU) and hateful class loss re-weighting (HW) are used (UNITER\textsubscript{CV10 + CFU + HW}), we observe further improvements on the development set. The Phase 1 results, however, show a larger increase in performance compared to UNITER, which indicates that these modifications help the model to generalize better. When increasing the number of folds to $15$ (UNITER\textsubscript{CV15 + CFU + HW}), we observe additional improvements on the Phase 1 test set; while $30$-fold cross-validation (UNITER\textsubscript{CV30 + CFU + HW}) gives an even further increase in dev-seen performance, this does not translate to improvement on the Phase 1 test set. 
%Due to these shortcomings, the model might receive too little information to improve upon our baselines. This could be resolved by training a stronger detection model than YOLO on fine-grained classes.
We also find that margin ranking loss (UNITER\textsubscript{CV15 + CFU + HW + MRL}) and YOLO9000 \cite{redmon-yolo9000} object tags (UNITER\textsubscript{CV15 + CFU + HW + YOLO}) do not improve Phase 1 performance. We notice that, for memes targeting black people and Muslims, YOLO9000 predicts a range of different labels such as Nigerian, Ugandan, or Ethiopian, potentially adding noise to the model. We furthermore find that it fails to detect objects such as guns, missiles and wheelchairs in several images. Such noise may therefore be responsible for the effects on performance. Some approach of fine-tuning YOLO on the HM dataset might help to improve results.

UNITER\textsubscript{ENSEMBLE 1} consists of predictions obtained as an EA-optimized ensemble over three model variants -- UNITER trained using 30-fold cross-validation, UNITER with margin ranking loss, and UNITER with YOLO9000 object tags. Each of these versions of UNITER are themselves results of ensembles of cross-validation style training on different folds (based on the HM training set) and optimized using EA as detailed in Section \ref{sec:crossval}. Since these models employ different strategies, we hypothesize their errors will be different and therefore ensembling may lead to complementary effects and help to improve performance. This ensemble achieves the highest Phase 1 score -- $79.13$. The second ensemble we evaluate is UNITER\textsubscript{ENSEMBLE 2} which includes the aforementioned three UNITER models as well as three UNITER models trained using 15-fold cross-validation on the training set and using varying sets of seeds and batch sizes. This model leads to an improvement in performance albeit a particularly small one. Finally, UNITER\textsubscript{FINAL} uses 15-fold cross-validation where part of the development set (dev-seen) is included in the per-fold training set, as described in Section~\ref{sec:crossval}. 
This set of three different ensemble models are the ones that we submitted to Phase 2 of the challenge (the final phase of the competition), with the last one, UNITER\textsubscript{FINAL}, achieving an AUROC of $80.53$ and ranking us $4$th on Phase 2's leaderboard. This improvement in performance could be attributed to the fact that \textit{dev-seen} has more truly multimodal instances, and incorporating these during training enables the model to reason better using both the modalities. 

\section{Error analysis and further model variants}
\label{error}

In this section, we present qualitative analyses we conducted for one of our best models by looking at its misclassification patterns on dev-seen, and detail additional experiments we designed in an attempt to improve performance further but which were not successful. We hope that these observations can further inform the basis of future research in the field. Specifically, we designed two more experiments to supplement the model with additional, complementary information, as well as experimented with independent attention blocks for different modalities to mitigate overfitting on the text features. The details of the experiments are presented below.

\subsection{Analyzing misclassifications} \label{sec:misclass}

\begin{figure}[t!]
     \centering
     \begin{subfigure}[b]{0.3\textwidth}
         \centering
         \includegraphics[width=\textwidth]{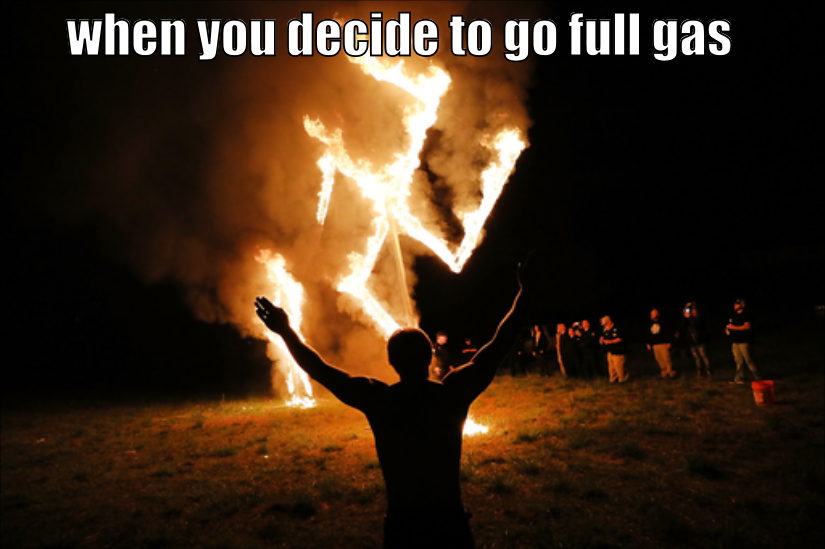}
         \caption{}
     \end{subfigure}
     \hspace{.4cm}
     \begin{subfigure}[b]{0.3\textwidth}
         \centering
         \includegraphics[width=\textwidth]{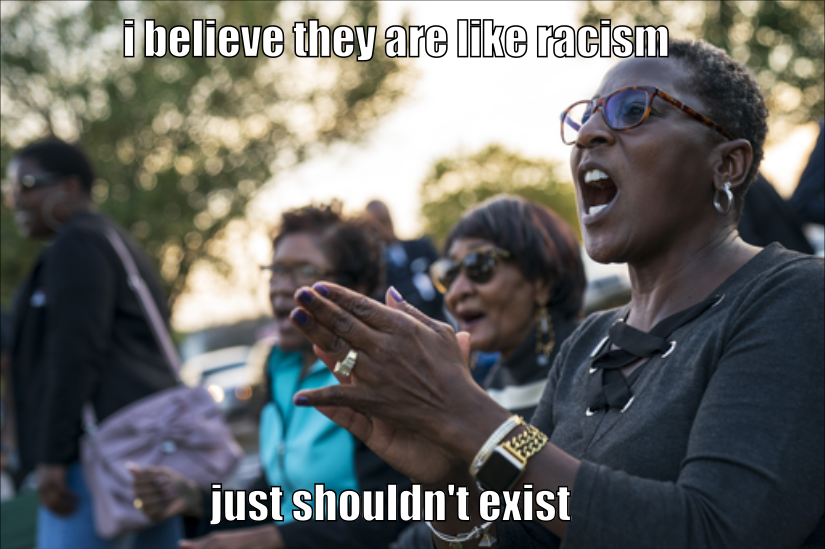}
         \caption{}
     \end{subfigure}
     \hspace{.4cm}
     \begin{subfigure}[b]{0.3\textwidth}
         \centering
         \includegraphics[width=\textwidth]{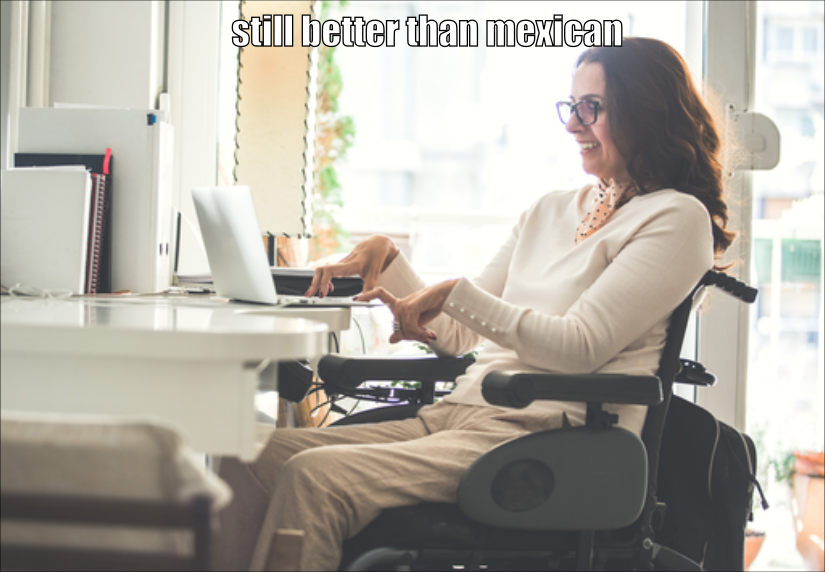}
         \caption{}
     \end{subfigure}
     
     %%%%%%%%%
     
     \begin{subfigure}[b]{0.3\textwidth}
         \centering
         \includegraphics[width=0.833\textwidth]{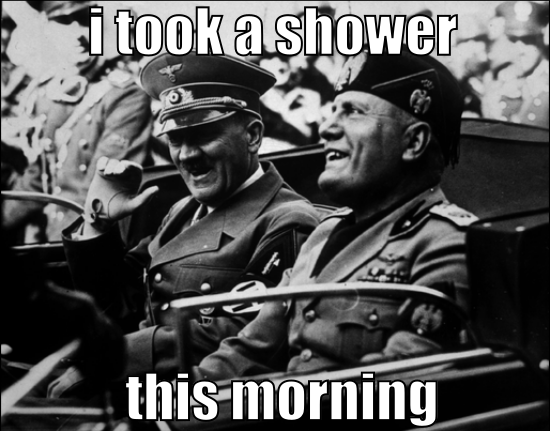}
         \caption{}
     \end{subfigure}
     \hspace{.4cm}
     \begin{subfigure}[b]{0.3\textwidth}
         \centering
         \includegraphics[width=\textwidth]{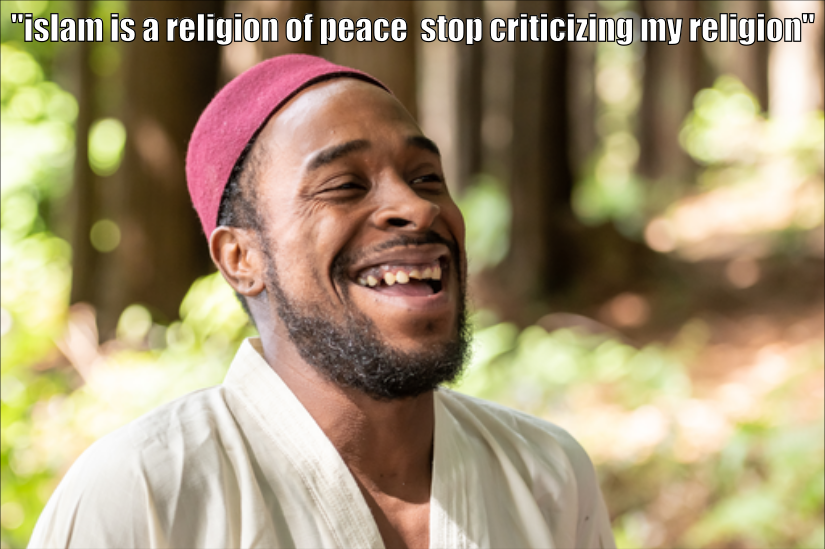}
         \caption{}
     \end{subfigure}
     \hspace{.4cm}
    %  \hfill
     \begin{subfigure}[b]{0.3\textwidth}
         \centering
         \includegraphics[width=\textwidth]{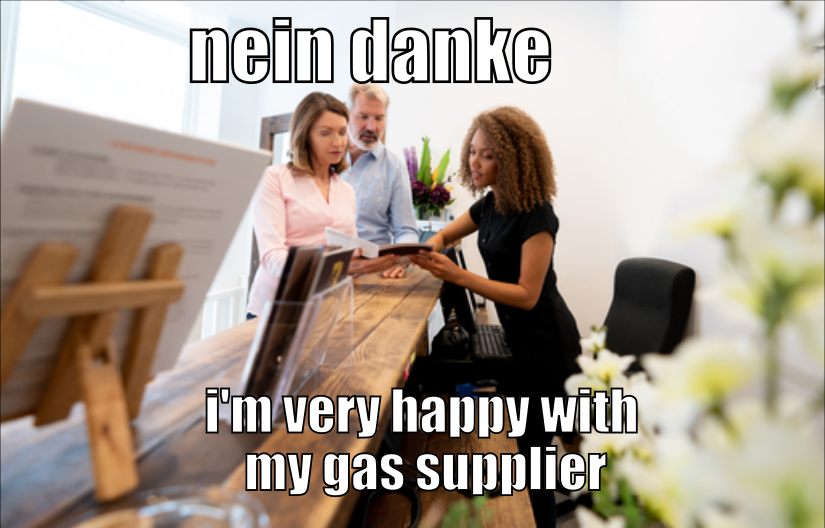}
        \caption{} 
     \end{subfigure}
     
    \caption{(a)-(c) False negatives: memes from the HM dataset that are labeled as hateful but predicted as being non-hateful. (d)-(f) False positives: memes that are labeled as non-hateful but predicted as being hateful. These are obtained from the UNITER\textsubscript{CV15 + CFU + HW} variant. Images above are a compilation of assets, including ©Getty Images. The hateful content does not represent the authors' views or opinions in any way whatsoever and it is only used for the purposes of demonstrating model misclassifications.}
    \label{fig:false_neg_pos}
\end{figure}

We investigated the misclassifications of UNITER\textsubscript{CV15 + CFU + HW} (one of the best models of Phase 1) to understand the errors the model makes and identify possible ways in which the model could be improved. We present some of the model's false negatives and false positives in Figure \ref{fig:false_neg_pos}. It can be seen that the false negative predictions are on memes that are truly multimodal and hateful in nature when image and text are put together. We find the model cannot reason about target groups; for example, Muslims, wheelchair users, and the Ku Klux Klan. Furthermore, it does not recognize real-life persons such as Anne Frank, or symbols. 

The false positives are also mainly multimodal as some of them could become hateful if a different text or image is used. Here too, the ability to identify the target groups might help the model to better identify hate and benign memes.

\subsection{Capturing target group information} \label{sec:minority}

In light of the observations above, we hypothesized that identifying %the minorities or groups of people targeted in the meme 
the target groups in the memes may help improve classification performance. Below we present another method aimed at including this information in the model. %as input to the model.

\paragraph{Social Bias Frames} 
We also performed experiments using the Social Bias Inference Corpus (SBIC) \citep{sap-socialbias}. This dataset contains around $150$k structured annotations regarding lewdness, offensiveness, intent to offend, targeted group and more across $44,761$ social media posts. Our first set of experiments focused on multi-task learning (MTL), i.e., simultaneously learning to classify memes as hateful/non-hateful and predicting SBIC classes as an auxiliary task. Specifically, we experimented with both offensiveness classification (binary) and target group prediction (multi-class) as the auxiliary task in separate experiments using the SBIC training data. During MTL training, we trained on alternate batches of the primary and auxiliary tasks with different classification heads for each. However, these setups yielded lower scores on dev-seen compared to simple fine-tuning of UNITER. % the dev-seen set of the hateful memes task. 
Subsequently, %to explicitly provide the model with information about these social groups, 
we experimented with an alternative implementation where we first fine-tune RoBERTa \citep{liu-roberta} solely on SBIC's target group classification task and then used the model to generate target group labels for each of the memes (based on the meme text as input). We then fed this as input to our UNITER model (similarly to the process used with the YOLO9000 dataset; Section \ref{yolo}). However, this too failed to improve results, possibly because of noise in the generated labels and the fact that the meme text alone is unlikely to be sufficient for accurate identification of the target group.  

\subsection{Emotion detection} 
\citet{rajamanickam2020joint} showed that multi-task learning with emotion detection can lead to significant improvements in abusive language detection. Taking inspiration from this work, we implemented a multi-task learning variant with emotion detection as the auxiliary task using the GoEmotions dataset \citep{demszky-goemotions}. The dataset consists of annotations for $58$k Reddit comments over $27$ emotion categories including a neutral class (multi-label). Similarly to SBIC, we trained on batches of the primary and auxiliary tasks sampled in different ratios, and with different classification heads for each task. However, the model lead to no improvements in performance on dev-seen.

\subsection{Overfitting text features} %Combating text features overfitting
During our experiments, we found that the model performs well on image confounders but struggles on text confounders. This suggests that it relies more on textual features and not as much on image features. In order to mitigate this, we split the internal attention layers of the UNITER transformer architecture into four independent chunks to enable varying dropout rates on different modalities. Specifically, the single attention mechanism was split into \textit{text-to-text, text-to-image, image-to-image} and \textit{image-to-text} blocks, allowing us to apply different dropout rates on each section. We also ensured that the dropout is consistent across all the multiple-attention heads of each layer, i.e., the dropout across the independent chunks was broadcasted along all the heads. We aimed for a higher \textit{text-to-text} dropout and lower \textit{image-to-text} and \textit{text-to-image} dropouts. We found that this approach also fails to improve results further.

\section{Conclusion}

We proposed a multimodal hateful memes detection framework that ranked 4th in Phase 2 of the 2020 Hateful Memes Challenge launched by Facebook. The proposed solution showcases a number of effective techniques: how to utilize the truly multimodal memes (confounders) in the dataset during training to strengthen the model's reasoning capability; how to effectively optimize an ensemble of models based on cross-validation and an evolutionary algorithm for weight tuning;  re-weighting the loss of the minority class; using image features in the correct feature space, i.e., using the exact same checkpoint of the object detector backbone as used by these multimodal architectures during their pretraining phase. We furthermore conducted an error analysis based on model misclassifications and detailed additional experiments that were designed in an attempt to improve performance further but that were not successful. We hope that these observations can further inform future research in the field. 

We note that for both our proposed solution and other state-of-the-art architectures such as Oscar there is a lot of room for improvement in terms of their ability to perform ``true'' multimodal reasoning. Fine-tuning the image extractor during training can help to improve image understanding with respect to the task. However, multimodal reasoning ability is hindered by sub-par image understanding largely due to lack of world knowledge in the current image feature extractor architectures. Identifying target groups %racial and religious features in people, people with disabilities, and other physical attributes in general 
as well as public figures and symbolisms are important aspects that can facilitate accurate interpretation of multimodal content.  %capturing their hateful correlations with the text of the meme. 
We leave the implementation of alternative approaches to capturing such information for future work. % exploring these avenues as interesting directions of future work.
%Furthermore, fine-tuning the image extractor during training. Updating it (or parts of it) during training could also potentially improve image understanding specific to this task. 
Recent developments in the field, such as ERNIE-ViL \citep{yu-ernie-vil}, a multimodal model that incorporates structured knowledge from scene graphs \citep{Johnson-scenegraphs} in its pretraining tasks, %and leads to better and more robust representations for downstream V+L tasks, 
is another interesting avenue for future work. 

%Using this as a base model would perhaps give further improvements on the hateful memes task. 

% \section*{Broader Impact}

% \ph{Might be nice to also have this section included, given all other NeurIPS papers had to do it.}

% \begin{ack}
% Acknowledgements \ph{Lisa cluster?}
% \end{ack}

\bibliographystyle{bibstyle}
\bibliography{neurips_2020}

\newpage
\appendix

\section{Appendix}
\subsection{YOLO9000 relevant classes} 
\label{sec:yolo_classes}
Table \ref{tab:yolo9000_classes} presents the list of the $97$ YOLO9000 classes we use in our experiments:

\begin{table}[ht!]
    \centering
    \small
    \caption{List of YOLO9000 classes used in our models.}
    \begin{tabular}{|c|c|c|c|c|}
    \toprule
    military soldier & marcher & domestic goat & domestic sheep & Nigerian \\
    Muslimah & Zimbabwean & Sudanese & Eritrean & Punjabi \\
    Yemeni & Ugandan & niqab & wild goat & sniper rifle \\
    Tommy gun & revolver & machine gun & Black African & mountain gorilla \\
    pygmy chimpanzee & gal & Mongol & Tibetan & farm boy \\
    cover girl & homeless & amputee & Guyanese & Iraqi \\
    heavyweight & Albanian & guy & Nicaraguan & Abyssinian \\
    South African & Cameroonian & Haitian & Jordanian & Afghan \\
    lady & old man & Ethiopian & Kenyan & lass \\
    Senegalese & clown & general & captain & minister \\
    ambassador & wheelchair & ram & missile & bomber \\
    goat herder & diocesan & eparchy & great grandson & Labrador retriever \\
    Korean & schoolchild & Lithuanian & Bolivian & Japanese \\
    Arabian & stallion & trotting horse & Omani & Ugandan \\
    Bornean & orphan & ape & sweetheart & waiter \\
    freight train & German shepherd & Siberian husky & Eskimo dog & hydrogen bomb \\
    khakis & Father & hijab & Guinean & Papuan \\
    monk & native & Kalashnikov & Mexican & stylist \\
    rabbi & beard & kitten & kitty & fireman \\
    man & Yugoslav  & & &  \\
    \bottomrule
    \end{tabular}
    
    \label{tab:yolo9000_classes}
\end{table}

\end{document}